\documentclass{article}
\usepackage{spconf,amsmath,graphicx}
\usepackage{todonotes}

\title{US-Net for robust and efficient nuclei instance segmentation}
\name{Zhaoyang Xu$^{\star }$ \thanks{* Correspondence to Zhaoyang Xu (zhaoyang.xu@qmul.ac.uk). The work described in this paper is partially supported by the EPSRC grant titled “Computer models for CRLM progression assessment based on histopathological image scans”, reference EP/N034708/1. 
This research utilised Queen Mary's Apocrita HPC facility, supported by QMUL Research-IT.}  \qquad Faranak Sobhani$^{\star }$  \qquad  Carlos Fern{\'a}ndez Moro$^{\dagger}$ \qquad Qianni Zhang$^{\star}$}

\address{$^{\star}$ Multimedia and Vision Group (MMV), Queen Mary University of London, United Kingdom.\\
    $^{\dagger}$ Dept. of Clinical Pathology/Cytology, Karolinska University Hospital, Stockholm, Sweden}

\begin{document}
%
\maketitle
\begin{abstract}
We present a novel neural network architecture, US-Net, for robust nuclei instance segmentation in histopathology images.
The proposed framework integrates the nuclei detection  and segmentation networks by sharing their outputs through the same foundation network, and thus enhancing the performance of both.
The detection network takes into account the high-level semantic cues with contextual information,
while the segmentation network focuses more on the low-level details like the edges.
Extensive experiments reveal that our proposed framework can strengthen the performance of both branch networks in an integrated architecture and outperforms most of the state-of-the-art nuclei detection and segmentation networks.

\end{abstract}
%


\section{Introduction}
\label{sec:intro}

One of the main challenges in manual traditional pathology evaluation based on H\&E (Haematoxylin and Eosin) stained slides, is the significant time, efforts and skills required for visual assessment of each case. A massive number of samples are being produced on a daily basis, requiring to be examined. Meanwhile, the increasing shortage of subspecialised pathologists is being reported. Fortunately, with the recent advances in digitization techniques for scanning digital whole slide images, a good foundation is laid for developing intelligent computer-aided histopathology assessment systems. Such systems are expected to augment the pathologists' ability by automating some fundamental, labor-intensive and relatively easy tasks, and allowing the experts to focus on the most challenging parts of the assessment. 
The analysis on cell shape, size, distribution, and other features is an essential task for both biologists and histopathologists in their visual analysis of histology data. Similarly, the automation of this task plays a critical role for subsequent analysis in computer-aided histopathology image assessment. 
The localisation and classification of cell types provide important clues in some disease diagnosis. For example, the spatial distribution of the cells can be utilised as unique features for tumour segmentation.


Recently, deep convolutional neural networks achieve impressive performance on object detection and segmentation tasks \cite{girshick_2016_region}, and they open new opportunities for tackling the challenge of automatic nuclei detection and segmentation in histology images. Dozens of successful deep learning based object detection and segmentation methods have been proposed, including two-stage object detection methods like Fast R-CNN\cite{girshick_2015_fast} and Faster R-CNN\cite{ren_2017_faster}. These methods cascade the features from two stages for better results.
While one stage methods like "single shot multibox detector" (SSD)\cite{liu_2016_ssd} and "you only look once" (YOLO) \cite{redmon_2016_you} are faster speed since all the procedures are being accomplished in one network while keeping a comparative accuracy against two-stage networks. 
Of all the segmentation networks, U-net\cite{ronneberger_2015_u} is prevalently used for biomedical image processing due to its concise and efficient structure comparing to other segmentation networks like deep lab \cite{chen_2018_deeplab}. 
However, these mentioned networks are either too powerful and complex or too simple and ineffective to directly produce decent nuclei detection results. Specially designed networks are required to address the unique nature of the data in nuclei detection and segmentation, like high-density, occlusion and limited range of shapes and sizes. 

\begin{figure*}[h]
\centering
\includegraphics[width=\linewidth]{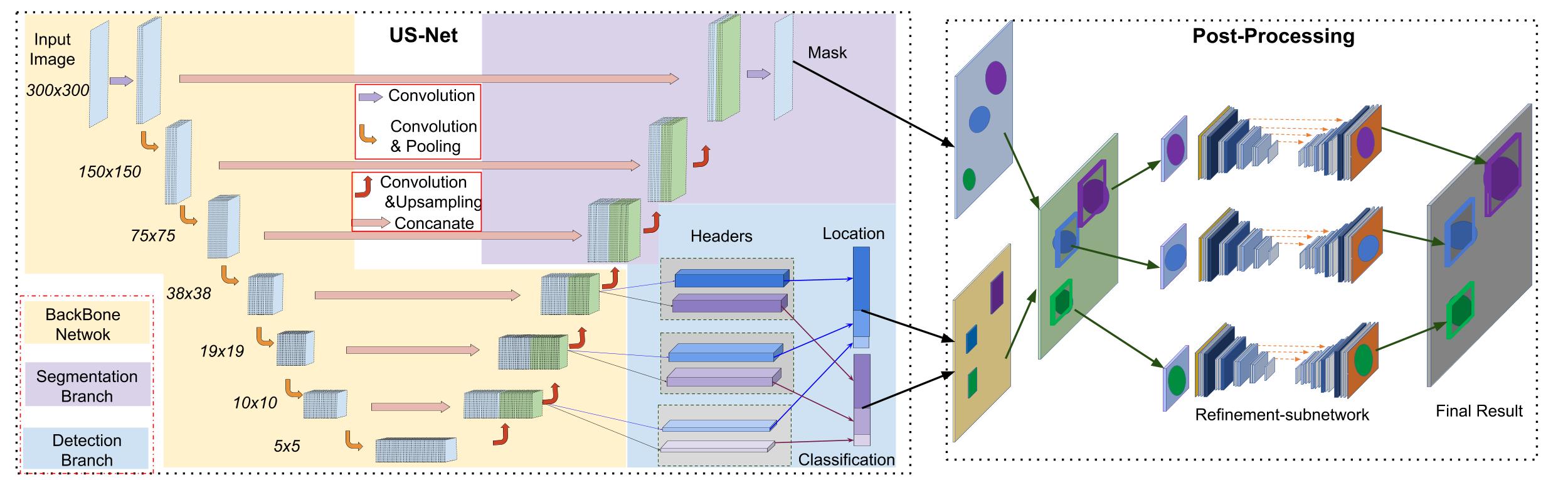}
\label{fig:us_net}
\vspace{-0.3cm}
\caption{US-Net with post-processing sub-network}
\end{figure*}

Comparing to natural objects, the detection and segmentation of nuclei seem to be much easier due to their simple structures and homogeneous properties in representation.  However, despite of the fact that the topic of nuclei detection and segmentation has been studied for decades, there is still no publicly available trained models that support universal nuclei detection across H\&E slides of different labs and conditions. Before the broadly adoption of deep neural networks, conventional nuclei detection methods often use the statistical or geographical features of images to generate the seeds.  In most cases, colour deconvolution is a necessary pre-processing or normalization step for guaranteeing a coherent performance on different datasets.

In the era of deep learning, various networks are proposed to solve the challenge. 
The work by Xie \textit{et al.}~\cite{xie2016microscopy}  proposes a fully convolutional regression network structure with good performance on the overlapping and clumping cells. Another regression network represented in~\cite{akram2016cell} employs the bounding box for cell (nucleus) detection.  
To address the common problem of lack of training data, other attempts take semi-supervised or unsupervised approaches to solve this task. Xu~\textit{et al.} try to extract the features of the nuclei with an unsupervised network, stacked sparse auto-encoder, and then use the extracted features to classify the foreground and background ~\cite{xu2016stacked}. Yet, the task of automating nuclei detection and segmentation remain under-addressed, due to various reasons including the lack of training sets, the high visual variance of data from different sources, etc.

Therefore in this paper, we aim to propose a robust model for nuclei detection and segmentation that can produce accurate and coherent results on independent H\&E image datasets with varying conditions.
This model, referred to as US-Net, benefits from a concise, yet efficient architecture, which consists of a nuclei detection network and a segmentation network. It involves a work flow that dynamically integrates the regression output of nuclei location and the end-to-end output of the semantic segmentation to enhance the performance of both networks.

The main contribution of this research is two-fold:
i) a novel and robust deep neural network architecture for instance segmentation of the nuclei in H\&E stained histopathology image; and
ii) an enhanced focal loss that can help deal with the class imbalance and accelerate the training is designed. 

\section{US-Net for nuclei detection and segmentation}
\label{sec:method}
To tackle the task of precise, instant and generic nuclei detection and segmentation in H\&E histology images, a specifically designed network architecture US-Net is proposed in this research. 
As shown in Fig.\ref{fig:us_net}, the structure of US-Net is very compact, composed of segmentation and detection branches, which share the same backbone network.
In principle, the propose network takes advantage of the powerful end-to-end semantic segmentation ability of U-net \cite{ronneberger_2015_u} structure and the excellent object detection and classification performance of SSD \cite{liu_2016_ssd}, precisely RetinaNet \cite{lin_2017_focal}, to achieve instance segmentation results with the help of a post-processing sub-network for refinement.

The overall objective of the network extends the loss for MultiBox objective \cite{erhan_2014_scalable}: given an input image $I$ and its corresponding segmentation masks  $s$ , location information $l$ and its class information $c$ , the loss $L$ can be decided by the following  function: 
\begin{equation}
L(I,c,l,s) = L_{conf}(I,c) +\alpha L_{loc}(I,l) +\beta L_{seg}(I,s)
\end{equation}
where the parameters $\alpha $ and  $\beta$  control the relative importance of the loss components.


The term $L_{seg}(I,s) $  which is defined with the $L1$ norm, helps to achieve the segmentation results. 

\begin{equation}
L_{seg}(I,s) =\dfrac{1}{M} \sum||s_{(x,y)}-\hat{s}_{(x,y)}||
\end{equation}
where $\hat{s}$ is the segmentation outcome of the segmentation branch and  $s$  is the ground truth information. $M$ is the total number of pixels  in the input image.
The loss $L_{conf}$ measures the confidence scores of the binary class (nuclei or not) of the detected boxes with an adapted version of Focal loss \cite{lin_2017_focal} : 

\begin{equation}
L_{conf}(I,c) =  \eta_{\hat{s}c}(1-\hat{c})^{\gamma}log(\hat{c})
\end{equation}
where $c \in {0,1}$  denotes the ground truth information for classification and $ \eta_{\hat{s}c}$ is associated with the output of the segmentation results:
\begin{equation}
\eta_{\hat{s}c} = \sqrt[]{\dfrac{S_{all}}{\hat{S_{c}}}}
\end{equation}
where $S_{all}$  represents the number of the pixels in ground truth $s$, while $\hat{S}_{c}$ denotes the number of pixels that equals to $c$ in output of the $\hat{s}$.
The term $L_{loc}(I,l)$ in Eq. \ref{eq:loc_loss} calculates the location regression loss of the multi-boxes given the ground truth information $l$ with a Smooth $L1$ loss as defined in \cite{erhan_2014_scalable}. 
\begin{equation}
L_{loc}(I,l) = \dfrac{1}{N}\sum_{i=1}^{N}\sum_{p\in (x,y,w,h)}^{}smooth_{L1}(l_{i}^{p}-\hat{l}_{i}^{p})
\label{eq:loc_loss}
\end{equation}
For the $i^{th}$ bounding box location information $l_{i}$,  $(x,y)$ is the center of the box while $h$ and $w$ represent the height and width of the box. 
With the Soomth $L1$ loss defined as:
\begin{equation}
smooth_{L1} = \begin{cases} 0.5x^{2} &  |x| < 1\\ |x| -0.5 & |x| \geq 1 \end{cases} 
\end{equation}
the output of US-Net can only achieve a relatively rough segmentation result. Thus, post-processing steps are necessary to realise an instance level segmentation of the input image. For this purpose, another  U-net structure network that consisting of 4 convolutional layers is built for further refining the detected regions.

\begin{figure*}[h]
\centering
\includegraphics[width=\linewidth]{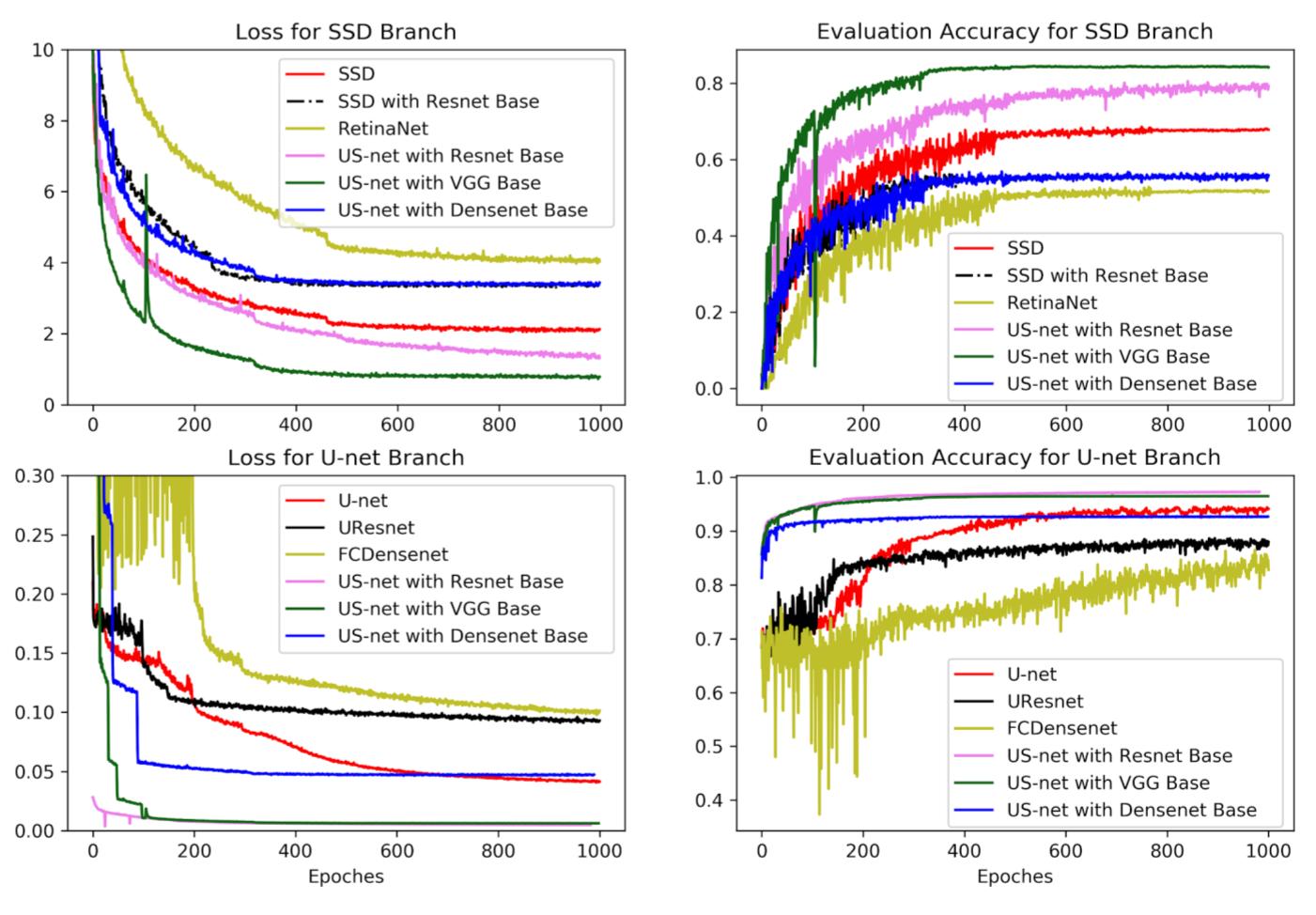}
\caption{Losses and evaluation accuracies for different network structures}
\label{fig:loss_curve}
\end{figure*}

\section{Experiments and Evaluation}
\subsection{Dataset}
The training dataset  employed in the experiments come from the Segmentation of Nuclei in Images Contest (SNIC)\footnote{“Digitalpathology:Segmentationofnu-cleiinimages.” [Online]. Available:http://miccai.cloudapp.net/competitions/83} and the MICCAI MoNuSeg \footnote{ “Mulit-organ nuclei segmentation challenge.” [Online]. Available: https://monuseg.grand-challenge.org/Home/}. 
There are 32 patches with size 600$\times$600 pixels from SNIC and  30 patches with size 1000$\times$1000 pixels from MoNuSeg. Both of them have instance level annotation.
The proposed model works with input patches of size 300$\times$ 300.  Hence,  the images from the original dataset are cropped to size 400$\times$400 with a fixed step size of 200.
After pre-processing, 878 patches are acquired in which 650 patches are used for training and 228 patches for evaluation.
For all the nuclei from different organs, they are treated as the same kind of nucleus which means no category information will be attached to each nucleus since the accurate detection and segmentation of the nuclei is the main focus.
In addition to the 300$\times$300 patches, another dataset with patch size 48$\times$48 is needed for the refinement sub-network in the post-processing stage. These patches are cropped from the scaled bounding boxes area and then resized to 48$\times$48.

\subsection{Implementing Details}
\label{sec: training}
In the U-net part, there are six down-sampling layers and six up-sampling layers connected by a bottleneck layer. The block size for all the layers is 4.
In the SSD part,  the detectable objects' (nuclei) size is constrained to the range of 20$\sim$128 by using the feature maps from the last three layers of the base network. The corresponding feature map sizes are $38\times38, 19\times19, 10\times10$.
Two anchor box aspect ratios ($1\times1$ and $1\times0.75$), and two scales (0.8 and 1.2) are considered in the experiments which make four different anchor boxes for each point in the feature maps. That makes up the  7620 default anchor boxes for an input image.

In the training phase, the parameter $\alpha$  is set to 1 while  $\beta$  is set to 0.1. 
Two different optimizers are employed for the two branches. For the SSD branch,  Adam optimizer \cite{kingma_2014_adam} with a learning rate of 0.001  is employed, while for the U-net branch, the optimizer is  SGD with a learning rate 0.0001, momentum 0.9 and weight decay 0.0001.

\subsection{Results}
The evaluation of the results of the proposed networks is divided  into two parts, the evaluation of semantic segmentation results and evaluation of detection results.
For the segmentation part,  pixel accuracy (PA) is calculated regarding the number of accurately predicted pixels out of the total pixels.
The evaluation metric for object detection part is the same as the interpolated average precision (AP) used for VOC dataset \cite{voc}.

\begin{figure*}[h]
\centering
\includegraphics[width=0.9\linewidth]{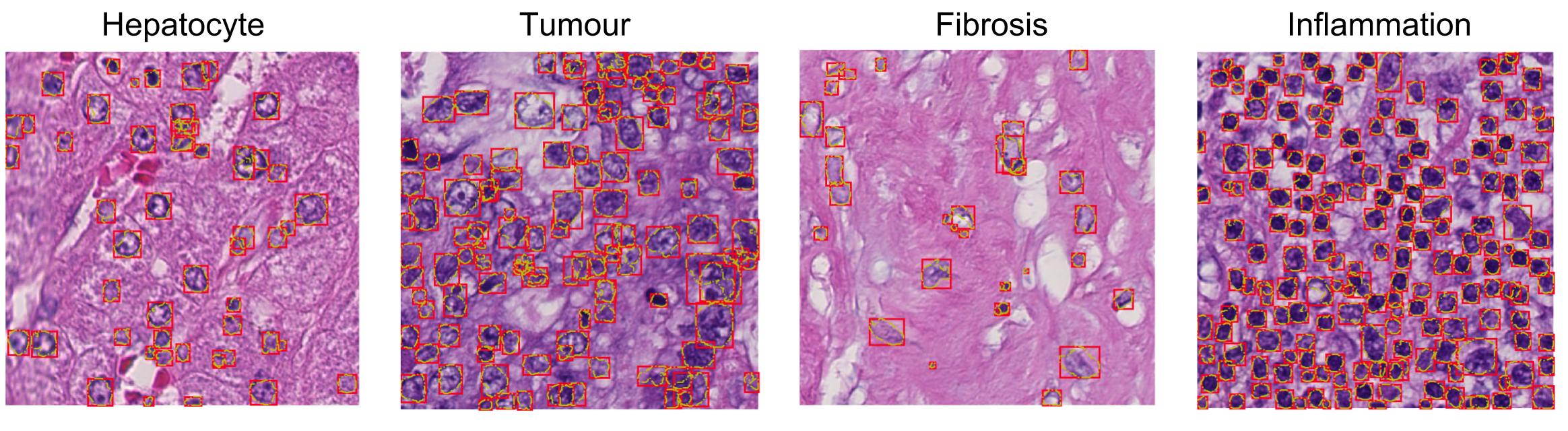}
\vspace{-0.5cm}
\caption{Results on the CRLM dataset}
\label{fig:crlm}
\end{figure*}

In Fig.\ref{fig:loss_curve},  losses for different branches of the US-Net as well as the AP/PA  for evaluation are demonstrated along the training process.
From the losses and AP/PA  in the curves, it can be observed that the US-Net performs much better than the any of the branch individually.
Besides, we can find that by adding the Densenet blocks or Resnet blocks to the networks, the performance is not necessarily imported due to the low complexity of the features in the image.
Furthermore, we apply the trained model to the histopathology images from colorectal liver metastasis patients.
From the results that demonstrated in Fig \ref{fig:crlm}, we can visually observe with the visual assessment that the trained network has decent robustness and transferability.

\section{Conclusion}
\label{conclu}
In summary, we have proposed a robust network architecture US-Net, for nuclei detection and segmentation.
The network incorporates the of the U-net and SSD networks together to realize a concise and powerful instance segmentation network for locating the nuclei in H\&E stained images.
Comparison with other start-of-art methods demonstrated the efficiency of the proposed network.

\bibliographystyle{IEEEtran}
\bibliography{refs}

\end{document}